*Research Article*

# Deep Learning Based Cyberbullying Detection in Bangla Language

**Sristy Shidul Nath[1], Razuan Karim[2] and Mahdi H. Miraz[3,4,5,*]**

[1]University of Science and Technology Chittagong, Bangladesh
sristynath54321@gmail.com
[2]American International University-Bangladesh, Dhaka, Bangladesh
rkarim@aiub.edu
[3]Xiamen University Malaysia, Sepang, Selangor, Malaysia
m.miraz@ieee.org
[4]Wrexham University, Wrexham, UK
m.miraz@ieee.org
[5]University of South Wales, Swansea, UK
m.miraz@ieee.org
*Correspondence: m.miraz@ieee.org

**Abstract:** The Internet is currently the largest platform for global communication including expressions of opinions, reviews, contents, images, videos and so forth. Moreover, social media has now become a very broad and highly engaging platform due to its immense popularity and swift adoption trend. Increased social networking, however, also has detrimental impacts on the society leading to a range of unwanted phenomena, such as online assault, intimidation, digital bullying, criminality and trolling. Hence, cyberbullying has become a pervasive and worrying problem that poses considerable psychological and emotional harm to the people, particularly amongst the teens and the young adults. In order to lessen its negative effects and provide victims with prompt support, a great deal of research to identify cyberbullying instances at various online platforms is emerging. In comparison to other languages, Bangla (also known as Bengali) has fewer research studies in this domain. This study demonstrates a deep learning strategy for identifying cyberbullying in Bengali, using a dataset of 12282 versatile comments from multiple social media sites. In this study, a two-layer bidirectional long short-term memory (Bi-LSTM) model has been built to identify cyberbullying, using a variety of optimisers as well as 5-fold cross validation. To evaluate the functionality and efficacy of the proposed system, rigorous assessment and validation procedures have been employed throughout the project. The results of this study reveals that the proposed model's accuracy, using momentum-based stochastic gradient descent (SGD) optimiser, is 94.46%. It also reflects a higher accuracy of 95.08% and a F1 score of 95.23% using Adam optimiser as well as a better accuracy of 94.31% in 5-fold cross validation.

## 1. Introduction

While various contemporary technologies, such as social media and similar other communication platforms, bring us closer to each other and let us stay connected, they pose multiple serious threats to our lives, including cyber bulling [1,2]. Since the adoption and usage of the social networking platforms has mushroomed in the recent past years, bullying is no longer limited to the physical space; rather, social media has brought into being a new model of this tyrannised behaviour, i.e., cyberbullying. The dissemination of unsuitable photos, videos, insulting statements or messages, threats, etc., using the cyberspace in any form, such as texts, images, voice, videos and so forth, is considered as cyberbullying. Cyberbullying is the term for bullying that occurs online, including on social media, messaging apps, over the voice call, gaming platforms and mobile devices[1]. It is a pattern of behaviour intended to frighten, infuriate or shame those who are the target [3-5]. These cyberbullying behaviours include threats, trolling, information sharing and transmission of offensive images or videos. In fact, negative news rapidly spreads and becomes viral much faster than any positive ones, resulting in significant negative impacts on our lives, including deaths and suicidal cases. Therefore, in this technological era, it is extremely important to find efficient approaches and develop rapidly functioning tools to identify, detect and stop the spread of such inconsiderate actions [6-8].

Social media platforms like Facebook, Instagram, WhatsApp, LinkedIn, Twitter, WeChat, etc. are the highly popular ones amongst various age groups and thus act as major source of cyberbullying. In fact, there could be various motivations behind cyberbullying, including but not limited to one's appearance, academic or professional achievements, racism, sexuality, financial status, religions, etc. According to the recent statistics published by Broadband Search, a significant amount of harassment and bullying, including 42% on Instagram, 37% on Facebook, 31% on Snapchat, 12% on WhatsApp, 10% on YouTube and 9% on Twitter, occur on social media platforms[2]. According to the Daily Star[3], cyberbullying is on the rise in Bangladesh, where girls and women make up 80% of the victims and many of those incidents had led to suicide attempts.

According to Statista[4], during COVID-19 lockdowns, people including kids, engaged with various online platforms for 20% additional time compared to that of the pre-pandemic period. Although the COVID-19 pandemic is now over, this has left a long-term effect on us, particularly on how we use online platforms to satisfy our needs with regards to recreation and communication. According to data revealed by the Peu Research Center[5], 71% U.S. parents are concerned about their children aged 11 or younger, as they are spending excessive time engaging with the devices. Besides, numerous social media platforms has started to reveal various types of vulgar contents which directly motivated people to involve in cyberbullying.

Therefore, identifying, detecting, blocking and removing such contents is a matter of grave concern and demands extensive continuous research to be conducted, adopting various emerging tools including Artificial Intelligence (AI) based automation technologies. The aim of this study is to develop a model with higher accuracy for detecting cyberbullying in Bengali language using the Bi-LSTM model.

## 2. Literature Review

The influence of information technology on online communication has resulted in both positive and negative outcomes. Cyberbullying contributed to emotional distress and in extreme cases, like suicide, remain a significant concern. This literature review aims to explore existing studies, identify research gaps and assess the potential solutions for broader and more effective mitigation of the challenges associated with online cyberbullying [9].

Iwendi *et al*. [10] developed an advanced deep learning system for identifying obscenities in social statements. Bidirectional Long Short-Term Memory (Bi-LSTM), Gated Recurrent Units (GRU), Long Short-Term Memory (LSTM) and Recurrent Neural Network (RNN) have been used in their work. However, Bi-LSTM scored the highest accuracy of 82.18% compared with that of the other models.

Balakrishnan *et al*. [11] put forward a concept to improve cyberbullying detection using twitter users' psychological features and various machine learning classifiers, such as Naïve Bayes, Random Forest and J48. Their research demonstrated an accuracy of 91.88% when users' sentiments and personalities were used. However, lack of availability of the users' sentiment information can reduce the accuracy of this approach.

Maity *et al*. [12] developed a multitask multimodal framework called 'MT-MM-Bert with VecMap', based on Bert and VecMap embedding techniques for Hindi-English dataset, they obtained 82.05% accuracy in cyberbullying detection, 77.87% in sentiment analysis task and 58.05% in emotion recognition task.

Kumar *et al*. [13] proposed a multi-input integrated learning using deep neural networks for cyberbullying in real time code-mix data. In their research, they had used different word embedding techniques such as GloVe for English, Fast-Text for Hindi. For English data feature extraction, capsule network dynamic routing and for Hinglish data Bi-LSTM has been used.

Das *et. al* [14] focused on a machine learning model based on encoder-decoder, a renowned NLP tool, was developed to classify users' comments of Facebook pages. Seven numerous types of hate speech were found in a sample of 7,425 comments. For estimating hate speech classifications, this work used attention mechanisms, LSTM and GRU-based decoders. Amongst those algorithms, the attention-based algorithm showed the highest accuracy of 77%.

A significant study using several ML and DL techniques including Linear SVC, Logistic Regression, Multinomial Naïve Bayes, RF, ANN, and RNN were analysed by Emon *et al*. [15]. The study showed deep learning based RNN outperforms and gained accuracy 82.20% had been achieved.

Ahmed *et al*. [16] developed two particular models CNN and MNB for three different datasets of 5000 Bangla text, 7000 Romanised Bangla text and combination of 12000 Bangla and Romanised text, respectively. However, CNN models achieved the best performance, using the separate datasets; an accuracy of 84% in the Romanised Bangla dataset and 80% was achieved using the combined dataset.

Ahammed *et al*. [17] developed two different models based on SVM and MNB with TF-IDF feature extraction techniques, having a lower dataset of 1339 comments. The accuracy achieved in Naïve Bayes was 72%, while SVM obtained an accuracy of 70%.

Ghosh *et al*. [18] proposed cyberbullying detection in Bengali language strategies using various machine learning models: support vector machine, logistic regression, random forest and passive aggressive classifiers where TF-IDF and bag of words embedding techniques were used. The dataset was labelled with 'bully' on the basis of sexual, threat, troll and religious contents or 'not-bully'. There were three other categories: to whom the comment passed, adjacent gender and number of reactions. An accuracy was 78.1% have been achieved through this model.

Ahmed *et al*. [19] developed a hybrid neural network model using binary classification and multiclass classification, having a dataset of 44,001 Facebook comments, classified bully comments having an accuracy of 85%. This model has some limitation of false detection in terms of large sentences.

Tripto *et. al* [20] developed deep learning based models to classify Bengali sentences having three class sentiment labels such as positive, negative and neutral as well as another five class sentiment labels including strongly positive, positive, neutral, negative and strongly negative. A model, to figure out the six different emotions that can be present in a Bengali sentence, i.e., rage, disgust, fear, joy, sadness and surprise, was created. In their research, a variety of YouTube videos, comments in Bangla, English and Romanised Bangla (Bangla in English) were used. Two different word embedding techniques, i.e., Continuous gag of words (CBOW) and skip gram (SG), have been used to preprocess the data. Both deep learning models, such as LSTM and CNN, as well as machine learning models, such as naïve bayes and SVM, were used. Their research reveals that the LSTM models based on deep learning outperform basic machine learning models, with an accuracy of 65.97% and 54.24% for 3 class and 5 class sentiments, respectively. In terms of emotion detection, accuracy of the LSTM models was 59.23%. It was also observed that amongst the deep learning, LSTM model which performed better than CNN.

Chakraborty *et. al* [21] developed multinomial naïve bayes, support vector machine, CNN-LSTM based approaches to detect cyberbullying based on Bengali Unicode text and Unicode emoticons, gathering dataset from different pages of Facebook. For feature extraction, TF-IDF has been used in MNB and SVM classifier. They have also used linear support vector classifier and radial basis function kernal in support vector classifier (RBF SVC). Their results revealed that amongst these three models, SVM with linear kernal performed better than others, demonstrating 78% accuracy.

Tuhin *et al*. [22] separately introduced two models using naïve bayes and topical approach for sentiment analysis of Bengali language and dealt with an emotion-based dataset consisting of happiness, sadness, tenderness, excitement, anger and scaredness. The dataset was created manually. Their topical approach, using TF-IDF feature extraction, performs better achieving a 90% accuracy.

Sultana *et al.* [23] proposed six numerous machine learning models including logistic regression, multinomial naïve bayes, random forest, support vector machine, K-nearest neighbour and gradient boosting. For word embedding, TF-IDF transformer and TF-IDF vectorizer were used. Their work obtained a better accuracy of 85.7% using SVM classifier.

Shah *et al.* [24] conducted a research work based on various ML models, such as SVM, KNN, logistic regression and RF, using a larger dataset of 15307 rows which was taken from English comments and a smaller dataset of 3000 rows which was taken from Hinglish (Hindi in English) comments. Their results showed an accuracy of 92% on average.

As social media platforms are open to multimodal data such as texts, images, videos, etc., cyberbullying on such platforms are not only limited to through the of texts, but also through images and videos. In this regard, Roy *et al.* [26] developed a CNN based approach with transfer learning model, to detect toxic images. They have managed to achieve an accuracy of 89% using a comparatively small dataset of 1339 images.

Raj *et al.* [27] proposed numerous ML models and neural networks techniques to detect bullying in English language. The dataset was collected from Wikipedia. A 5-fold-cross validation technique have been used for training and testing every models. Different word embedding techniques such as count vectorisation, TF-IDF, GloVe and paragram were used with various ML models, i.e., naïve bayes, XG boost, SVM and logistic regression. Moreover, multiple neural networks, such as CNN, LSTM, GRU, Bi-LSTM, Bi-GRU, CNN with BiLSTM and attention based Bidirectional LSTM were also used with those word embedding techniques. According to their findings, for this purpose, the machine learning methods weren't as effective the as neural network-based models. Amongst the ML models, SVM with TF-IDF word embedding performed comparatively better with a 95.02% accuracy. Amongst the neural-network based techniques, Bi-GRU with GloVe embedding technique outperformed and the other models achieving an accuracy of 96.98%.

Sultan *et al.* [28] analysed performance of deep learning as well as machine learning based models for text detection, using three distinct datasets including Twitter data, offensive language and cyberbullying data. Their research showed the efficacy of the text detection techniques using deep learning model based, such as CNN, LSTM and Bi-LSTM, are better than that of machine learning based models, such as, NB, KNN, SVM, RF, DT with BoW, TF-IDF and DL Algorithms. In each dataset analysis, the Bi-LSTM based model outperformed others. The accuracies of all the machine learning models were below 90%.

Reghunathan *et al.* [29] developed three different machine learning models to compare the accuracy using two different languages, viz. English and Malayalam (one of the Indian native languages). Using TF-IDF techniques in feature extraction, they have obtained accuracies of 90% and 93.75%, using SVM for English and Malayalam dataset, respectively. Their results showed that SVM outperformed other models, such as logistic regression and random forest.

Wu *et al.* [30] conducted a study on the causes of cyberbullying and proposed a model to detect factors of cyberbullying. To conduct large data analysis for two instances of cyberbullying, they suggested a model based on the BERT classifier. They identified some of the aspects of cyberbullying and developed a manual annotation corpus. This work has collected many comments and posts from social media or forums and distributed in six classes or types of cyberbullying.

There are also some other studies using Bi_LSTM model, to detect the cyberbullying. Bilal *et al.* [31] have proposed various neural networks-based techniques to detect cyberbullying in Roman Urdu language. They have collected a dataset of 30000 comments from Facebook and Twitter, where half of the comments (i.e., 15000) were hate-speeches and remaining half of the comments were neutral. They have developed an annotation guideline for Roman Urdu hate speech and context-aware based model, employing CNN, LSTM, Bi-LSTM and attention based Bi-LSTM approaches. The attention based Bi-LSTM demonstrated the highest accuracy of 87.50%. However, their plans for future work include detecting sentiment of a sentence. That being said, the quality of the annotation of the Roman Urdu hate speech dataset may help increasing the accuracy. Kumar *et al.* [32] proposed a multi-input integrated learning using deep neural networks for cyberbullying in real time code-mix data. In their research, they had used different word embedding techniques, such as GloVe for English and Fast-Text for Hindi. For English data feature extraction, capsule network dynamic routing and for Hinglish data Bi-LSTM have been used. Both the models achieved an accuracy of 97%.

The summary of the previous researcher works has been provided in Table 1:

**Table 1.** Comparison of the Proposed Model with Other Existing Different Deep Learning Models

| Author | Dataset | Methodology | Feature Extraction | Result | Limitations |
|---|---|---|---|---|---|
| Das *et.al* (2021) [14] | 7425 | LSTM, GRU | TF-IDF | 77% accuracy | Imbalanced dataset where 6020 are hate speech |
| Emon *et al.* (2019) [15] | 4700 | Linear SVC, MNB, RF, ANN, RNN with LSTM | Count Vectorizer, TF-IDF | highest accuracy of 82.20% in RNN-LSTM | Lower amount of data where 90% training data, 10% testing data |
| Md. T. Ahmed *et al.* (2021) [16] | 5000 Bangla and 7000 Romanized Bangla | CNN, Multinomial Naïve Bayes | TF-IDF | For Bangla, accuracy of 84% in CNN, 80% in MNB for Romanize Bangla | Lower dataset |
| Ahammed *et al.* (2019) [17] | 1339 (665 hate, 674 normal) | SVM, Naïve Bayes | Count Vectorizer, TF-IDF | 72% accuracy in NB, 70% accuracy in SVM | Lower accuracy and lower dataset |
| R. Ghosh *et al.* (2021) [18] | -- | SVM, LR, RF, Passive Aggressive classifiers. | TF-IDF, BoW | 78.1% accuracy in PA with N-gram. | Lower accuracy |
| F. Ahmed *et al.* (N/A) [19] | 44001 FB comments (non-bully, sexual, threat, troll, religious) | CNN-LSTM, Ensemble Model | Word2vec | 87.9% accuracy in BC using CNN-LSTM and 85% accuracy in MC with Ensemble model | Higher complexity in operation |
| Tripto *et. al* (2018) [20] | 15689 (5011 Bangla, 4189 English, 6489 Romanized) | LSTM, CNN, NB, SVM | CBOW and Skip Gram | LSTM performs better. 3 class sentiment: 65.97%, 5 class sentiment: 54.24%, 59.23% in emotion | Lower accuracy due to low amount of data in 3 class, 5 class and emotion. |
| Chakraborty *et. al* (2019) [21] | 5644 (2739 abusive, 2905 non abusive) | Multinomial Naïve Bayes, SVM, CNN-LSTM | TF-IDF | SVM with Linear Kernal performs better with accuracy 78% | Lower dataset and accuracy |
| Tuhin *et al.* (2019) [22] | 7500, six types (happy, sad, tender, excited, angry, scared) | Naïve Bayes, Topical Approach | TF-IDF | Topical Approach with TF-IDF performs better with accuracy 90% | Lower dataset, Train: Test = 98: 2 |
| Sultana *et al.* (2023) [23] | 5000 | LR, SVM, KNN, RF, MNB | TF-IDF | Better accuracy in SVM with 87.5% | Lower dataset and accuracy |
| Shah *et al.* (2022) [24] | 15307 English and 3000 Hinglish | ML: SVM, KNN, LR, RF, Bagging, SGD, AdaBoost, MNB | Count vectorizer, TF-IDF | Around 92% accuracy on average | Imbalanced dataset (64.2% toxic data) |
| Atoum *et al.* (2020) [25] | 5628 Tweets (1187 positive, 2342 negative, 2099 neutral) | NB, SVM | N-gram, Chi square, Information gain | 92.02% in 4-gram | Lower dataset |
| Raj *et al.* (2021) [27] | 115,864 (13,590 toxic) | ML: XG Boost, NB, SVM, LR DL: CNN, LSTM, GRU, Bi-GRU, Bi-LSTM, Bi-LSTM- | CV, TF-IDF word unigram, bigram, trigram, and character bigram, trigram, GloVe, Paragram | 96.98% in Bi-GRU, 98.69% in Attention-BiLSTM with GloVe | -- |

| | | CNN, Att-BiLSTM | | | |
|---|---|---|---|---|---|
| Sultan *et al.* (2023) [28] | -- | ML: NB, KNN, SVM, RF, DT DL: | BoW, TF-IDF | Bi-LSTM performs better others with accuracy of 90.2% | -- |
| Reghunathan *et. al* (2022) [29] | English from GitHub and Malayalam | SVM, LR, RF | TF-IDF | 90% in English, 93.75% in Malayalam using SVM | -- |
| Wu *et al.* (2022) [30] | 125 posts and 5926 comments | Two different BERT models, SVM, RF | TF-IDF in SVM, RF | BERT models perform better over ML algorithms | Lower accuracy, Lower dataset |

## 3. Research Methodology

Figure 1 shows the proposed deep learning-based model for detection of cyberbullying in Bengali language, which consist of three fundamental steps:

1. Data Preprocessing
2. Model Development
3. Performance Evaluation

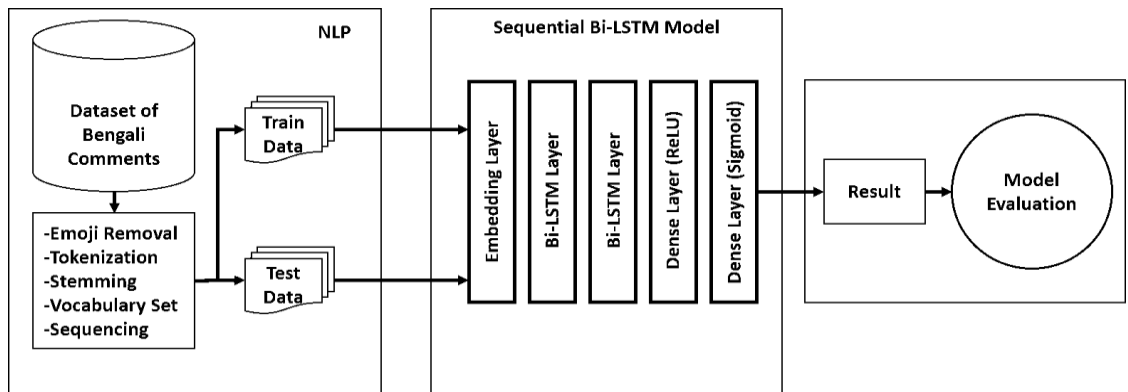

**Figure 1.** Proposed Deep Learning based Model for Detection of Cyberbullying in Bangla Language

### 3.1. Data Collection

As not much research on cyberbullying detection in Bengali language has been conducted thus far, dataset collection for this research was a very challenging task. To facilitate this research, an appropriate dataset has been produced by combining different datasets from Kaggle, Mendeley and some manually added ones. This has been done to ensure that the dataset used for this research is strong and versatile enough. In this project, as demonstrated in Figure 2, a balanced dataset of total 12282 social media comments in Bengali language have been used.

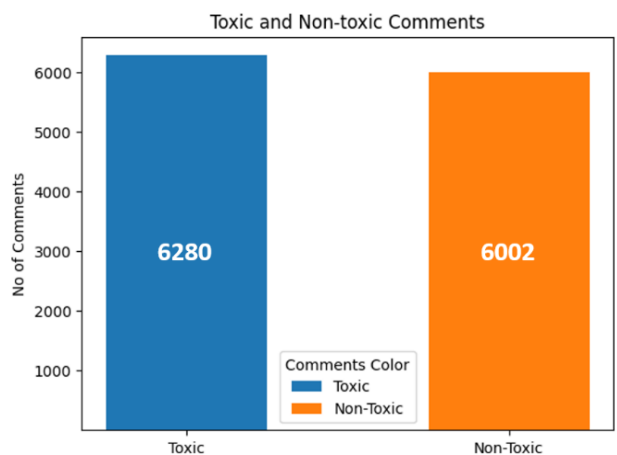

**Figure 2.** Data Classification

### 3.2. Data Preprocessing

The dataset was imported from 'csv' file and comments with adjacent level of toxic or non-toxic have been inserted into two distinct lists. This work has divided the comment section into four categories like threat, obscene, insult and racism. However, if the comment belongs to any of the mentioned categories, it will be considered as toxic otherwise as nontoxic. The dataset contained several emojis, as most of the phrases were taken from various internet platforms. Upon collection of the initial entries for the dataset, the data pre-processing steps have been completed, including emoji reduction, tokenisation, word stemming, vocabulary set construction and pad sequencing.

#### 3.2.1. Emoji or Data Cleaning

Before passing through the model, for training and testing purposes, the dataset was cleaned by removing the emojis as it could have a negative impact on the model's performance.

#### 3.2.2. Tokenisation

The technique of breaking down a text into smaller word chunks is known as tokenisation. In this research, the comments from the dataset were split into words.

#### 3.2.3. Stemming or Lemmatisation

The process in which the root of a word can be obtained is known as stemming or lemmatisation. For example: 'mature' is the root word of 'immature'. Therefore, in this step, the suffix and the prefix of a word has been eliminated for finding its root. This would also help detecting similar words, i.e., originated from the same root, in the future.

#### 3.2.4. Vocabulary Set Formation

This step has been carried out to reduce redundancy of the words and to produce a unique set of words.

#### 3.2.5. Sequencing

By using the indices of corresponding words from vocabulary set, the sentences have been reconstructed. For this purpose, the comments or sentences have been replaced with vectors of integers. To reduce complexity arising from the varied length of data, the sentences have been converted to a fixed length of 80, using pad sequence of 0 to the left of the vectors. This is because, in most of the cases, the toxic words are located either in the middle of the sentences or at end of the smaller sentences. However, completion of sequencing with padding thus turns the input vector shape into 12282×80. Upon successful accomplishment of the aforementioned steps (i.e., 3.2.1 – 3.2.5), the data pre-processing is completed, and the vector of the dataset becomes ready to be passed through the model for the training and testing purposes.

### 3.3. Data Split

The dataset was divided into two distinct portions with a ratio of 80 to 20. The ratio indicates while 80% of the dataset were utilised to train the model, the remaining 20% of the dataset were used to evaluate the model's performance.

### 3.4. Model Architecture

Figure 3 shows the basic architecture of proposed model which consists of an embedding layer, two Bidirectional LSTM layers and two dense layers. To simplify the things and get better features, this work applied pre-padding with a maximum length of 80 and a value of 0, which resulted in the reduction of longer sentences with lengths or word counts above 80.  In this model, an embedding layer is particularly a hidden layer that converts input data or information from a high-dimensional space to a low dimensional space, enabling the network to better understand how inputs relate to one another and process the data more promptly. The input dataset has been passed through the embedding layer and configured to turn every word into a vector with a size of 200, because it was used to demonstrate how similar words were, and the outcome was (9825 × 80) to (9825 × 80 × 200). The output of the embedding layers has been considered as an input of the Bi-LSTM layer. In this model, two Bi-LSTM layers (i.e., 64 neurons and 32 neurons) have been employed. Dense layers are generally employed to alter the dimensionality of the outputs from the Bi-LSTM layers. In this model, two dense layers have been added following the Bi-LSTM layers. The first dense layer of 64 neurons has been used to inactive the neurons of the previous layer to reduce the complexity. For that purpose, ReLU activation function has been used. The second dense layer of single neuron has been used to convert the output between (0, 1), because of binary classification problem. In this regard, the last dense layer has been employed with the Sigmoid activation function. The experiment was executed on 100 epochs and 1024 batch size.

### 3.5. Hyperparameter Tuning for Optimisation

The parameters that are controlled by the user, to regulate the model during the training or the learning stage, are referred as hyperparameters of machine learning.

#### 3.5.1. Learning Rate

The rate at which the model will progress through the learning stage has a profound impact by the learning rate. Furthermore, the user can alter the rate of learning based on error during the learning phase, in order to lessen it. The higher rate can directly affect the learning procedure to find an optimum value and it is quite challenging. However, it generates error and the system becomes unstable. Moreover, lower rate can reduce the speed of learning and it can be time consuming too.

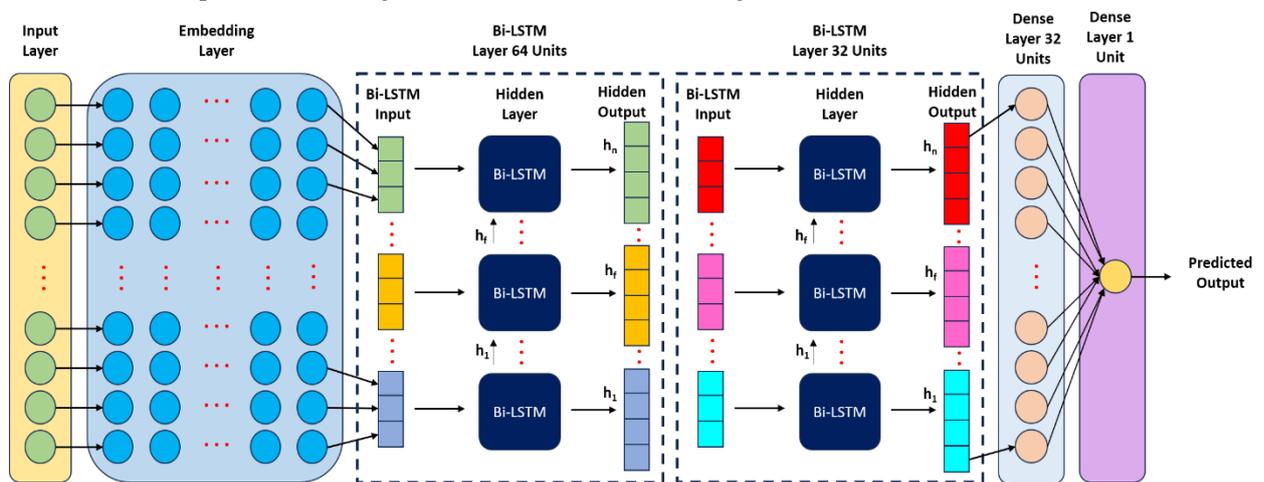

**Figure 3.** Basic Architecture of the Proposed Model

#### 3.5.2. Batch Size and Epoch

The training set separated into different subsets, which influence how rapidly a model learns and generalises by updating the model parameters, is referred as the batch size. Additionally, the parameter epoch shows how many times the model has been trained on the training data. Until the reduction in validation error, epochs can be increased. The epoch indicates when to stop learning if no improvement is obtained in validation error.

#### 3.5.3. Activation Function

By mapping the weighted sum and bias, the activation function regulates the output and lowers the complexity of the neurons. By deactivating neurons that have little impact on output, it establishes nonlinearity. Despite having numerous activation functions, two distinct functions were used in this research work.

#### 3.5.4. Optimiser

Algorithms that are used to minimise error and raise efficiency by controlling learning rate and weights are referred as the optimiser. By performing successive modifications to the parameters of each of the layers that are connected, the model performance is optimised.

### 3.6. Performance Evaluation and Mathematical Equation

The performance of the model can be obtained by calculating True Positive (TP), True Negatives (TN), False Positives (FP), and False Positives Negatives (FN). From those values, confusion matrices were also plotted. The developed classifier's performance is measured using a variety of performance indicators. In text categorisation, several typical performance metrics are examined in relation to the following metrics:

### 3.6.1. Accuracy

The accuracy of a model represents the number of correctly classified or predicted comments (number of true positives and number of true negatives). It can be defined as:

$$\text{Accuracy} = \frac{TP+TN}{TP+TN+FP+FN} \tag{1}$$

### 3.6.2. Precision

Precision score emphasises the precision of the model's successful predictions. Out of all positive occurrences that were predicted, it determines the proportion of correctly projected positive outcomes, as follows:

$$\text{Precision} = \frac{TP}{TP+FP} \tag{2}$$

### 3.6.3. Recall

Quantifying recall is an estimate of all positive occurrences that were accurately forecast. The higher the recall is, the more efficacy of predicting positive events is. Recall can be defined as:

$$\text{Recall} = \frac{TP}{TP+FN} \tag{3}$$

### 3.6.4. F-score

To provide an authoritative evaluation of the model's accuracy, F-score merges the precision score with the recall score into a single rating, as follows:

$$F-\text{measure} = \frac{2\times \text{Precision}\times \text{Recall}}{\text{Precision}+\text{Recall}} \tag{4}$$

## 4. Experiment and Result

To optimise the loss in this model, the Adam optimiser has been selected as default. This study analysed the model's performance by changing various model parameters such as batch size, epochs, initial learning rate, etc. Learning rate scheduler was used with this optimiser to lessen overfitting issue that have been found with SGD optimiser. By lowering the learning rate in accordance with a predetermined schedule, it can modify the learning rate during training. The model trained with an initial learning rate of 0.1 and the learning rate scheduler's constant decay rate was obtained by calculating the quotient of learning rate divided by the number of epochs. So, by altering different values from 0.5 to 0.9 of the momentum, the performance of the model was examined. SGD with momentum 0.9 and epoch at 50 could slightly solve the overfitting issue of using SGD optimiser. The model achieved a 99% training accuracy with 94.71% validation accuracy. The accuracy and F1 score of the model were 94.46% and 94.64%, respectively. Figure 4(a) and Figure 4(b) describe the resultant loss and accuracy curve per epoch during model compilation and Figure 5 illustrates the adjacent confusion matrix. Adam optimiser was employed to analyse the performance by changing model parameters such batch size, epochs and initial learning rate. From the analysis, better results were obtained with the initial learning rate of 0.00001, using 100 epochs. The loss and accuracy for training and validation data curves are plotted in Figure 6(a) and Figure 6(b), respectively. The outcome of the model's confusion matrix is also demonstrated in Figure 7. Both the training and validation accuracy levels were 99% and 95%, respectively. This model had a 95.23% F1 score and a testing accuracy of 95.08%.

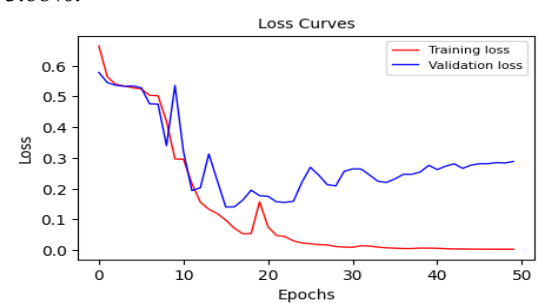

**Figure 4(a).** Training Loss and Validation Loss

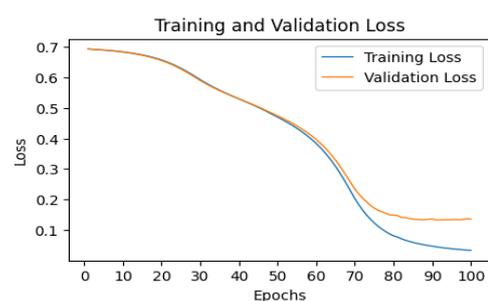

**Figure 4(b)**. Training Accuracy and Validation Accuracy Curve

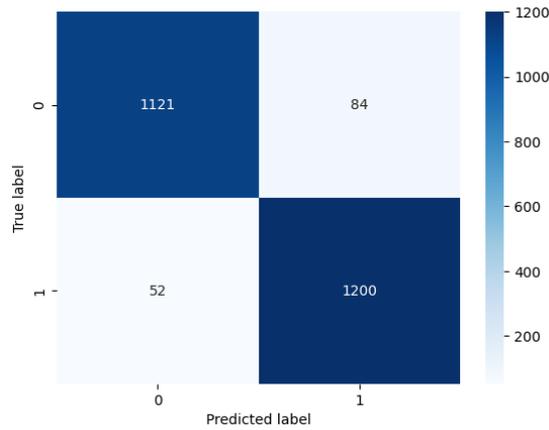

**Figure 5.** Confusion Matrix

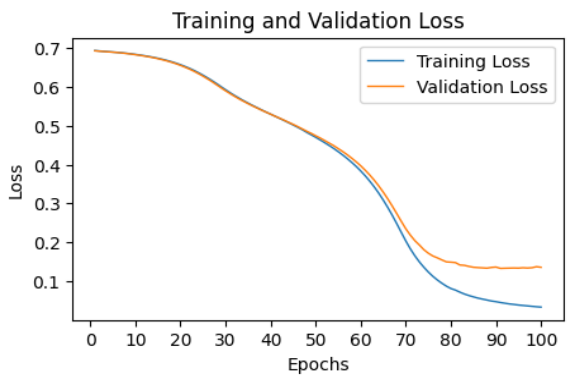

**Figure 6(a).** Training Loss and Validation Loss Curve

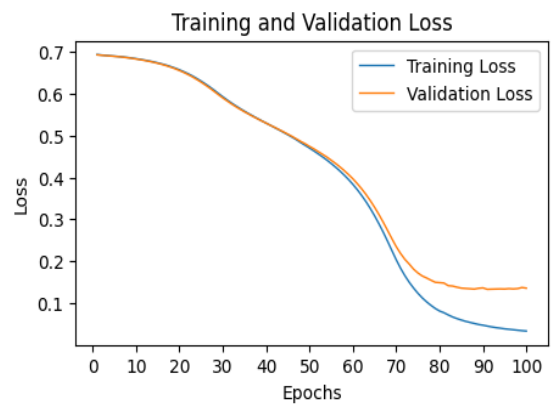

**Figure 6(b).** Training Accuracy and Validation Accuracy Curve

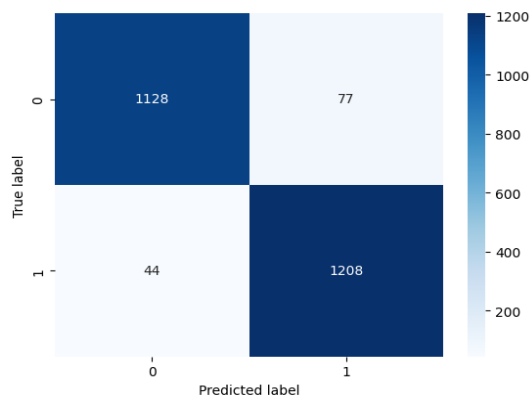

**Figure 7.** Confusion Matrix

The accuracy and loss curves for each fold are shown in Figure 8(a) and Figure 8(b); Figure 9(a), Figure 9(b); Figure 10(a) and Figure 10(b); Figure 11(a), Figure 11(b), Figure 12(a) and Figure 12(b), respectively. It is observed from the curves that by the repeated training and scoring the model on various segments of the dataset, k-fold cross-validation delivers a more accurate estimation of the model's efficacy than a single train-test split. It provides an evaluation metric that is more trustworthy and lessens the effect of data variability. A 5-fold cross validation was accomplished in order to verify the model's effectiveness. Adam optimiser is used, with epochs fixed at 100 per fold, for reducing the loss. By tuning hyper parameters several times, the best result was obtained in the 5-fold cross validation with a learning rate of 0.00001. The resultant average accuracy of this method was 94.31%. Figure 13(a), Figure 13(b), Figure 13(c), Figure 13(d) and Figure 13(e) exhibit the confusion matrices for the 5-fold cross validation.

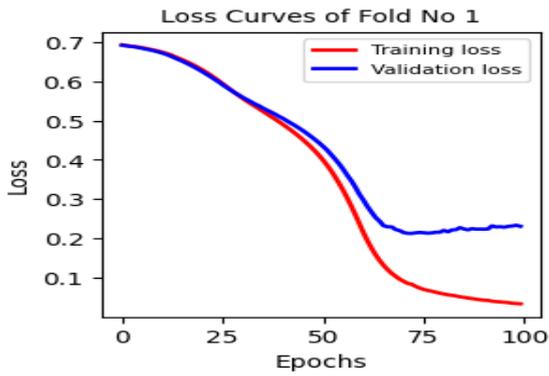

**Figure 8(a).** Training Loss and Validation Loss Curve

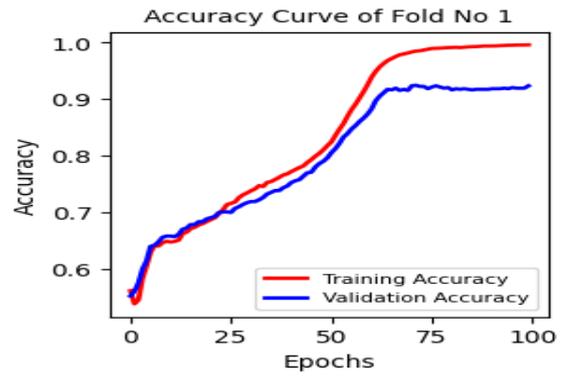

**Figure 8(b).** Training Accuracy and Validation Accuracy Curve of Fold 1

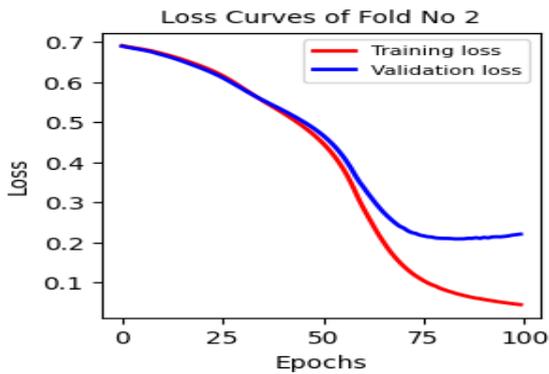

**Figure 9(a).** Training Loss and Validation Loss Curve

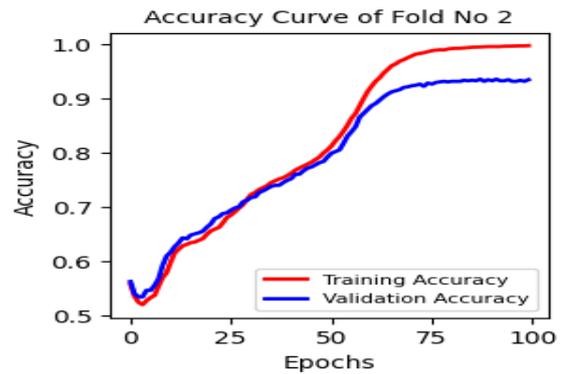

**Figure 9(b).** Training Accuracy and Validation Accuracy Curve of Fold 2

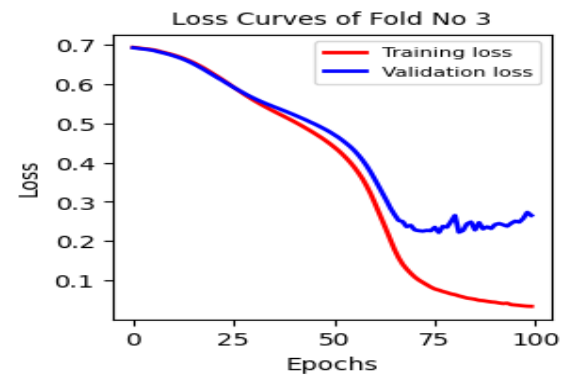

**Figure 10(a).** Training Loss and Validation Loss Curve

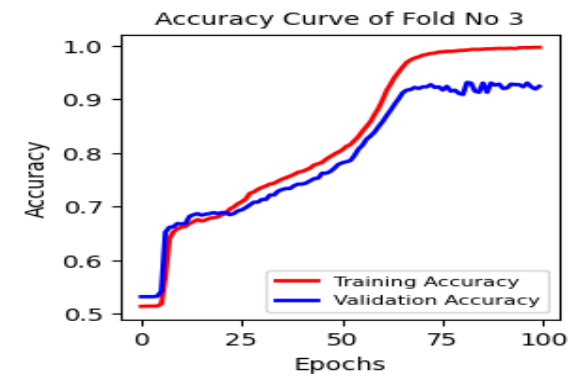

**Figure 10(b).** Training Accuracy and Validation Accuracy Curve of Fold 3

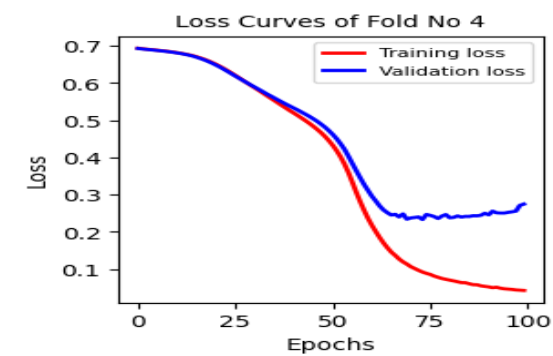

**Figure 11(a).** Training Loss and Validation Loss Curve

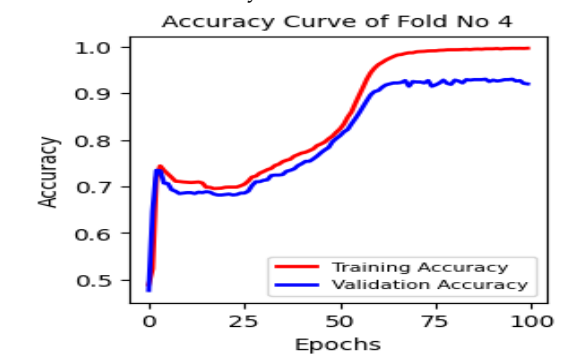

**Figure 11(b).** Training Accuracy and Validation Accuracy Curve of Fold 4

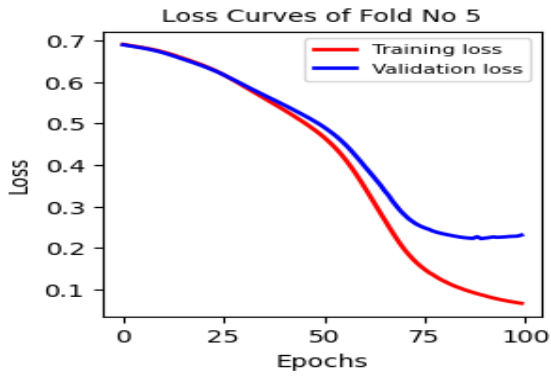

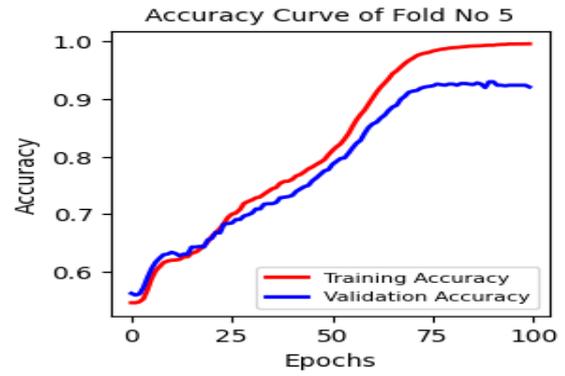

**Figure 12(a).** Training Loss and Validation Loss Curve

**Figure 12(b).** Training Accuracy and Validation Accuracy Curve of Fold 5

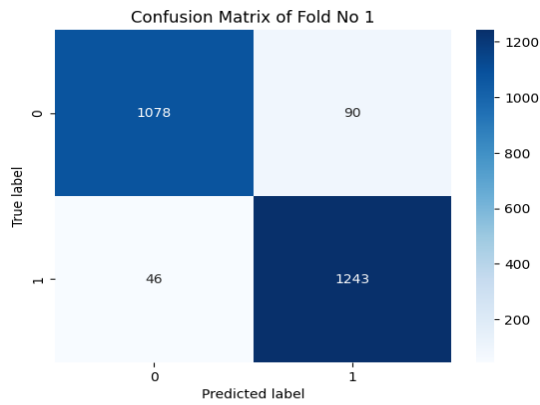

**Figure 13(a).** Confusion Matrix of Fold 1

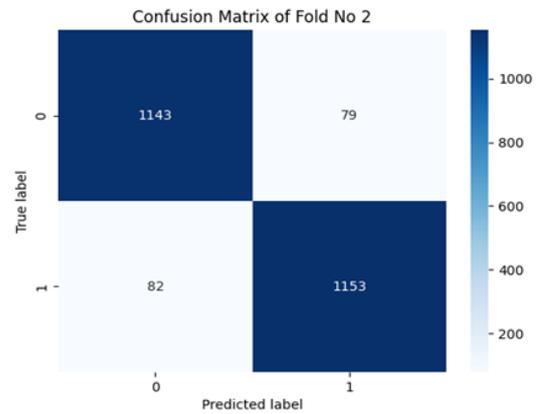

**Figure 13(b).** Confusion Matrix of Fold 2

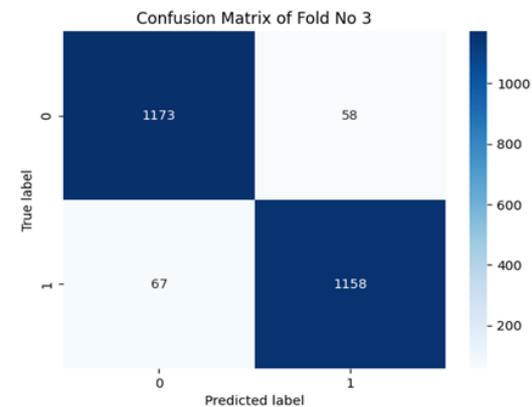

**Figure 13(c).** Confusion Matrix of Fold 3

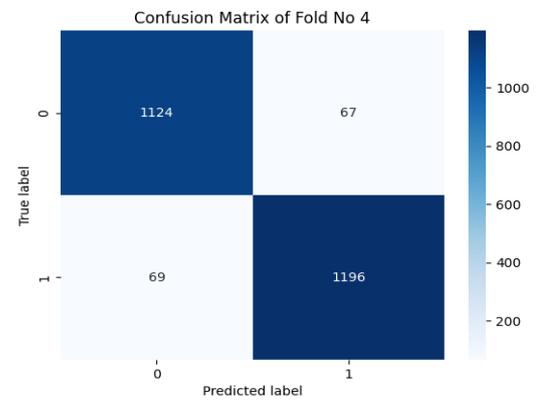

**Figure 13(d).** Confusion Matrix of Fold 4

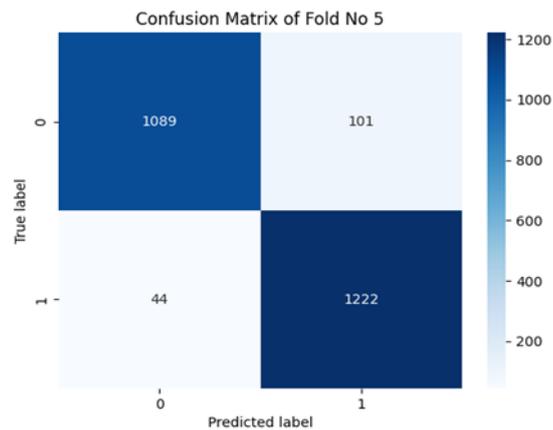

**Figure 13(e).** Confusion Matrix of Fold 5

Table 2 provides a summary of the corresponding performance results. From this table, it has been observed that f the proposed model has been achieved and verified using Bi-LSTM with momentum SGD, Bi-LSTM with ADAM optimisers and Bi-LSTM with 5-Fold Cross Validation technique. From all of the technique, the best performance has been verified by using 5-fold cross validation method using ADAM optimizer.

**Table 2.** Summary of the Corresponding Performance Results

| Method | Accuracy | F1 Score | Precision | Recall |
|---|---|---|---|---|
| **Bi-LSTM with momentum SGD** | 0.9446 | 0.9463 | 0.9345 | 0.9584 |
| **Bi-LSTM with Adam Optimiser** | 0.9508 | 0.9523 | 0.9400 | 0.9648 |
| **Bi-LSTM with 5-Fold Cross Validation** | 0.9446 | 0.9481 | 0.9324 | 0.9643 |
| | 0.9345 | 0.9347 | 0.9358 | 0.9336 |
| | 0.9491 | 0.9488 | 0.9523 | 0.9453 |
| | 0.9446 | 0.9462 | 0.9470 | 0.9455 |
| | 0.9410 | 0.9440 | 0.9237 | 0.9652 |
| Average Accuracy | | 0.9428 (±0.4883) | | |

The proposed model outperforms previous existing models with accuracy over 94% and overcomes some of the limitations, including lack of using higher dataset and low accuracy. Table 3 shows the comparison of the proposed model with other existing different deep learning models for cyberbullying detection in Bangla. It is noticed that the better accuracy of 95.08% was found from using Adam optimiser.

**Table 3.** Comparison of the Proposed Model with Other Existing Different Deep Learning models.

| Research Paper | Year | Dataset | Proposed Model | Result |
|---|---|---|---|---|
| **Das *et al.* [14]** | 2021 | 7425 | LSTM, GRU | 77% accuracy |
| **Ahmed *et al.* [16]** | 2021 | 5000 | CNN | 84% accuracy |
| **Present work** | 2023 | 12282 | 2 layers Bi-LSTM | 95.08% accuracy |

## 5. Limitation

This research work has concentrated on deep learning to determine whether a statement is toxic or nontoxic for Bangla language. The most challenging part of this research was to collect a dataset having adequate varieties as well as balancing and optimising the two layers Bi-LSTM model. To fit the model's SGD with momentum, Adam optimisers were separately used. Moreover, a 5-fold cross validation with Adam optimiser was used to analyse the model's performances. In all the cases, accuracy was above 90%. However, some noisy and anomalous sentences, such as 'তোরমতন ফাজিলের সাথে কথাবলাই উচিৎনা', 'অপদার্থকোথাকার' might have reduced the performance of the model. As in the first sentence the word 'তোরমতন', 'কথাবলাই', 'উচিৎনা' are not single words, rather these are combination of two different words i.e., 'তোর' and মতন', 'কথা' and 'বলাই', as well as 'উচিৎ' and 'না'. Similarly, in the second sentence, 'অপদার্থকোথাকার' is not a single word, it is rather a sentence comprising two different words, i.e., 'অপদার্থ' and 'কোথাকার'. Therefore, in such cases, the model couldn't learn the real meaning of sentences, resulting in a lower performance. On the contrary, in some cases, due to lack of variations in sentence structures, the model might have gained overfitting.

## 6. Conclusion and Future Work

As the prime goal of this project is to detect cyberbullying based on binary classification, the model was designed with different optimisers and to generalise the model's performance, a 5-fold cross validation was applied. The accuracy of the model with Adam optimiser is 95.08%, however, an overfitting issue was observed. To address overfitting problems, the learning rate scheduler is applied. Using the momentum-based SGD optimiser, the overfitting problem was minimised and an accuracy of 94.46% was achieved. Using a 5-fold cross validation, the accuracy was 94.31%. Therefore, it is anticipated that the model performed well on different variations of the datasets. Future research directions include further improving the model for multiclass classification as well as to address the limitations. Furthermore, deployment of the ensemble model, using numerous deep learning algorithms, to enrich the model's accuracy can also be considered. Attention-based model can also be deployed and evaluated for this purpose. TF-IDF and word2vec models, in word embedding techniques instead of default word embedding layer, could also be utilised to slightly raise the accuracy.

**Acknowledgement**

This research is financially supported by Xiamen University Malaysia (Project codes: XMUMRF/2021-C8/IECE/0025 and XMUMRF/2022-C10/IECE/0043).